\documentclass[10pt,twocolumn,letterpaper]{article}

\usepackage{cvpr}
\usepackage{times}
\usepackage{epsfig}
\usepackage{graphicx}
\usepackage{amsmath}
\usepackage{amssymb}
\usepackage{multirow}
\usepackage{bbm}
\usepackage{amsmath,bm}
\usepackage{tabularx}

\usepackage[breaklinks=true,bookmarks=false]{hyperref}
\newcommand{\tabincell}[2]{\begin{tabular}{@{}#1@{}}#2\end{tabular}}
\cvprfinalcopy 

\def\cvprPaperID{****} 

\begin{document}

\title{RTM3D: Real-time Monocular 3D Detection from Object Keypoints for Autonomous Driving}

\author{Peixuan Li$^1$ $^2$ $^3$ $^4$ $^5$, Huaici Zhao$^1$ $^2$ $^4$ $^5$,Pengfei Liu$^1$ $^2$ $^3$ $^4$ $^5$,Feidao Cao$^1$ $^2$ $^3$ $^4$ $^5$\\
$^1$Shenyang Institute of Automation, Chinese Academy of Sciences \\
$^2$Institutes for Robotics and Intelligent Manufacturing, Chinese Academy of Sciences\\
$^3$University of Chinese Academy of Sciences\\
$^4$Key Laboratory of Opto-Electronic Information Processing, Chinese Academy of Sciences\\
$^5$Key Lab of Image Understanding and Computer Vision, Liaoning Province\\
{\tt\small \{lipeixuan, hczhao,  liupengfei, caofeidao\}@sia.cn}
}

\maketitle

  \begin{abstract}
In this work, we propose an efficient and accurate monocular 3D detection framework in single shot. Most successful 3D detectors take the projection constraint from the 3D bounding box to the 2D box as an important component. Four edges of a 2D box provide only four constraints and the performance deteriorates dramatically with the small error of the 2D detector. Different from these approaches, our method predicts the nine perspective keypoints of a 3D bounding box in image space, and then utilize the geometric relationship of 3D and 2D perspectives to recover the dimension, location, and orientation in 3D space. In this method, the properties of the object can be predicted stably even when the estimation of keypoints is very noisy, which enables us to obtain fast detection speed with a small architecture. Training our method only uses the 3D properties of the object without the need for external networks or supervision data. Our method is the first real-time system for monocular image  3D detection while achieves state-of-the-art performance on the KITTI benchmark. Code will be released at \href{https://github.com/Banconxuan/RTM3D}{https://github.com/Banconxuan/RTM3D}.
  \end{abstract}
    \section{Introduction}\label{sec:introduction}
      \begin{figure}
		\begin{center}
			\includegraphics[width=1\columnwidth]{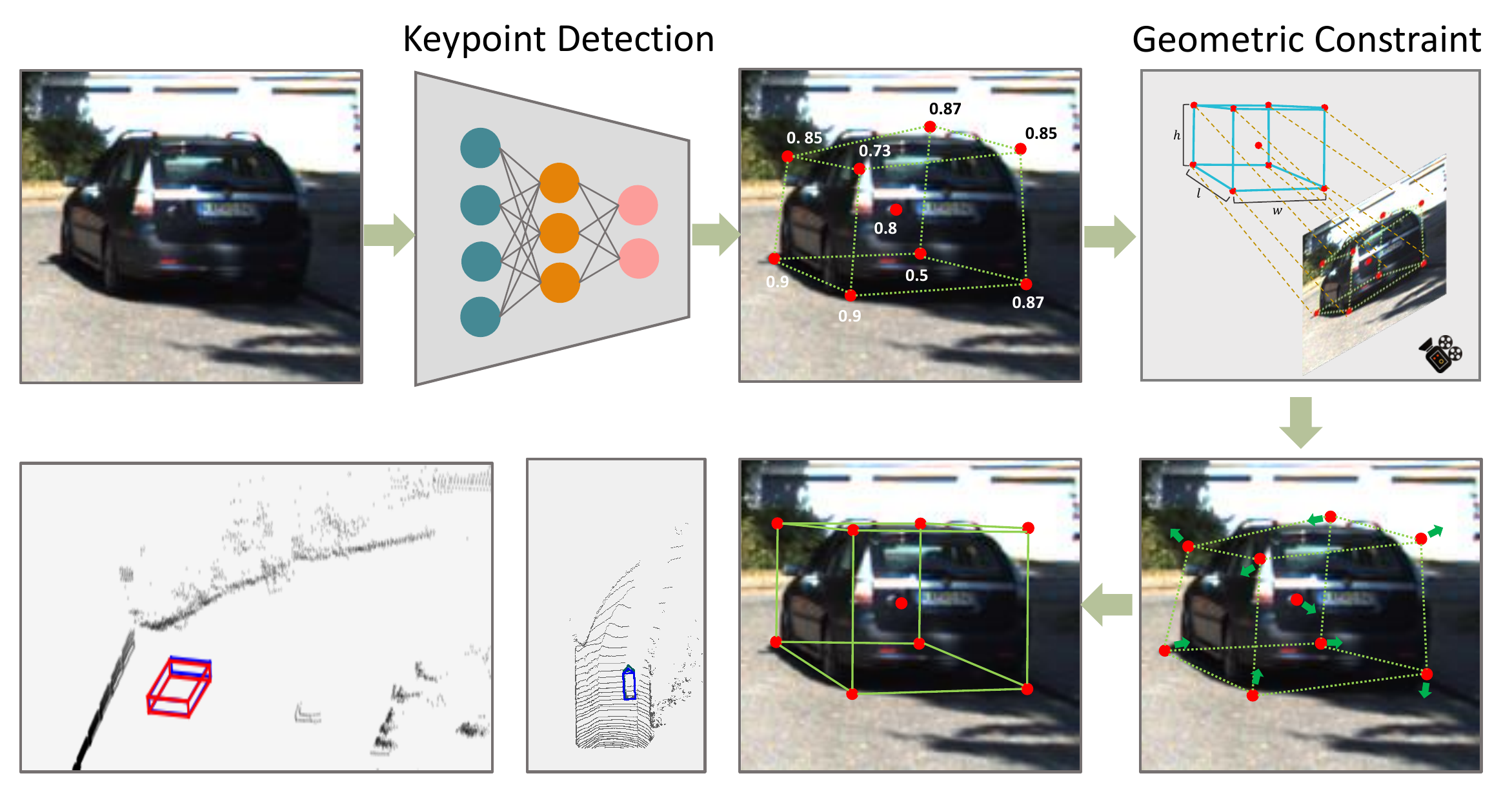}
		\end{center}
		 \caption{\textbf{Overview of proposed method:} We first predict ordinal keypoints projected in the image space by eight vertexes and a central point of a 3D object. We then reformulate the estimation of the 3D bounding box as the problem of minimizing the energy function by using geometric constraints of perspective projection.   }
		\label{fig:framework}
	\end{figure}
    3D object detection is an essential component of scene perception and motion prediction in autonomous driving \cite{behl2017bounding, geiger2012we}. Currently, most powerful 3D detectors heavily rely on 3D LIDAR laser scanners for the reason that it can provide scene locations \cite{chen2017multi,zhou2018voxelnet,yang2018pixor,qi2018frustum}. However, the LiDAR-based systems are expensive and not conducive to embedding into the current vehicle shape. In comparison, monocular camera devices are cheaper and convenient which makes it drawing an increasing attention in many application scenarios \cite{chen2016monocular,mousavian20173d,xu2018multi}. In this paper, the scope of our research lies in 3D object detection from only monocular RGB image.

    Monocular 3D object detection methods can be roughly divided into two categories by the type of training data: one utilizes complex features, such as instance segmentation, vehicle shape prior and even depth map to select best proposals in multi-stage fusion module \cite{chen2016monocular,chen20173d,xu2018multi}. These features require additional annotation work to train some stand-alone networks which will consume plenty of computing resources in the training and inferring stages. Another one only employs 2D bounding box and properties of a 3D object as the supervised data \cite{rubino20173d,brazil2019m3drpn,li2019stereo,yang2019cubeslam}. In this case, an intuitional idea is to building a deep regression network to predict directly 3D information of the object. This can cause performance bottlenecks due to the large search space. For this reason, recent works have clearly pointed out that apply geometric constraints from 3D box vertexes to 2D box edges to refine or directly predict object parameters \cite{Naiden2019ShiftRD,liu2019deep,brazil2019m3drpn,li2019stereo,mousavian20173d}. However, four edges of a 2D bounding box provide only four constraints on recovering a 3D bounding box while each vertex of a 3D bounding box might correspond to any edges in the 2D box, which will takes 4,096 of the same calculations to get one result \cite{mousavian20173d}. Meanwhile, the strong reliance on the 2D box causes a sharp decline in 3D detection performance when predictions of 2D detectors even have a slight error. Therefore, most of these methods take advantage of two-stage detectors \cite{girshick2014rich,girshick2015fast,ren2015faster} to ensure the accuracy of 2D box prediction, which limit the upper-bound of the detection speed.
\begin{table*}[t]
\setlength{\tabcolsep}{1mm}
\begin{center}
\begin{tabular}{|c| c  | c | c | c | c |}
\hline
Method&                             Real Time  &Stereo     &Depth     &Shape/CAD    &Segmentation\\
\hline
Mono3D \cite{chen2016monocular} &             &           &          &             & \checkmark\\
\hline
\tabincell{c}{3DOP \cite{chen20173d}, stereoRCNN \cite{li2019stereo}} &                      &\checkmark &          &             &            \\
\hline
\tabincell{c}{MF3D \cite{xu2018multi}, Peseudo-LiDAR \cite{wang2019pseudo},\\
 MonoPSR \cite{ku2019monocular} AM3D\cite{ma2019accurate}}&                     &           &\checkmark&             &            \\
\hline
Mono3D++ \cite{he2019mono3d++},Deep-MENTA \cite{chabot2017deep}, 3DVP \cite{xiang2015data} &              &           &&\checkmark&            \\
\hline
\tabincell{c}{
Deep3DBox \cite{mousavian20173d},GS3D \cite{Li_2019_CVPR},MonoGRNet \cite{qin2019monogrnet},\\
FQNet\cite{liu2019deep}, M3D-RPN\cite{brazil2019m3drpn} Shift-RCNN \cite{Naiden2019ShiftRD}
}&            &           &          &             &            \\
\hline
Ours(RTM3D)                                &\checkmark&           &          &             &            \\
\hline
\end{tabular}
\vspace{1mm}
\caption{Comparison of the real-time status and the requirements of additional data in different image-based detection approaches.}
\label{tab:extra}
\end{center}
\end{table*}

    In this paper, we propose an efficient and accurate monocular 3D detection framework in the form of one-stage, which be tailored for 3D detection without relying on 2D detectors. The framework can be divided into two main parts, as shown in Fig. \ref{fig:framework}. First, we perform a one-stage fully convolutional architecture to predict 9 of the 2D keypoints which are projected points from 8 vertexes and central point of 3D bounding box. This 9 keypoints provides 18 geometric constrains on the 3D bounding box. Inspired by CenterNet \cite{zhou2019objects}, we model the relationship between the eight vertexes and the central point to solve the keypoints grouping and the vertexes order problem. The SIFT, SUFT and other traditional keypoint detection methods \cite{lowe2004distinctive,bay2006surf}computed an image pyramid to solve the scale-invariant problem. A similar strategy was used by CenterNet as a post-processing step to further improve detection accuracy, which slows the inference speed.
    Note that the Feature Pyramid Network(FPN) \cite{lin2017feature} in 2D object detection is not applicable to the network of keypoint detection, because adjacent keypoints may overlap in the case of small-scale prediction. We propose a novel multi-scale pyramid of keypoint detection to generate a scale-space response. The final activate map of keypoints can be obtained by means of the soft-weighted pyramid.
    Given the 9 projected points, the next step is to minimize the reprojection error over the perspective of 3D points that parameterized by the location, dimension, and orientation of the object. We formulate the reprojection error as the form of multivariate equations in $\mathfrak{se}_3$ space, which can generate the detection results accurately and efficiently.
    We also discuss the effect of different prior information on our keypoint-based method, such as dimension, orientation, and distance. The prerequisite for obtaining this information is not to add too much computation so as not to affect the final detection speed. We model these priors and reprojection error term into an overall energy function in order to further improve 3D estimation.

    To summarize, our main contributions are the following:
    \begin{itemize}
		\item We formulate the monocular 3D detection as the keypoint detection problem and combine the geometric constrains to generate properties of 3D objects more efficiently and accurately.
		\item We propose a novel one-stage and multi-scale network for 3D keypoint detection which provide the accurate project points for multi-scale object.
        \item We propose an overall energy function that can jointly optimize the prior and 3D object information.
        \item Evaluation on the KITTI benchmark, We are the first real-time 3D detection method using only images and achieves better accuracy under the same running time in comparing other competitors.
	\end{itemize}
	\section{Related Work}\label{sec:related work}
    The 3D detection can be divided into two groups by the type of data: LiDAR-, and image-based methods.\\
    \textbf{LiDAR-based method.}
    LiDAR-based systems can provide accuracy and reliable point cloud of object surfaces in 3D scene. Therefor, most of the recent 3D object detection employ it in different representation to obtain the state-of-the-art model \cite{zhou2018voxelnet,beltran2018birdnet,chen2017multi,shi2019pointrcnn,li2016vehicle}.\\
    \textbf{Extra Data or Network for Image-based 3D Object Detection.}
    In the last years, many studies develop the 3D detection in an image-based method for the reason that camera devices are more convenient and much cheaper.
    To complement the lacked depth information in image-based detection, most of the previous approaches heavily relied on the stand-alone network or additional labeling data, such as instance segmentation, stereo, wire-frame model, CAD prior , and depth, as shown in Table. \ref{tab:extra}.
    Among them, monocular 3D detection is a more challenging task due to the difficulty of obtaining reliable 3D information from a single image. One of the first examples \cite{chen2016monocular} enumerate a multitude of 3D proposals from pre-defined space where the objects may appear as the geometrical heuristics. Then it takes the other complex prior, such as shape, instance segmentation, contextual feature, to filter out dreadful proposals and scoring them by a classifier. To make up for the lack of depth, \cite{xu2018multi} embed a pre-trained stand-alone module to estimate the disparity and 3D
    point cloud. The disparity map concatenates the front view representation to help the 2D proposal network and the 3D detection can be boosted by fusing the extracted feature after RoI pooling and point cloud. As a followup, \cite{ma2019accurate} combines the 2D detector and monocular depth estimation model to obtain the 2D box and corresponding point cloud. The final 3D box can be obtained by the regression of PointNet \cite{qi2017pointnet} after the aggregation of the image feature and 3D point information through attention mechanism, which achieves the best performance in the monocular image. Intuitively, these methods would certainly increase the accuracy of the detection, but the additional network and annotated data would lead to more computation and labor-intensive work.\\
    \textbf{Image-only in Monocular 3D Object Detection.} Recent works have tried to fully explore the potency of RGB images for 3D detection. Most of them include geometric constraints and 2D detectors to explicitly describe the 3D information of the object. \cite{mousavian20173d} uses CNN to estimate the dimension and orientation extracted feature from the 2D box, then it proposes to obtain the location of an object by using the geometric constraints of the perspective relationship between 3D points and 2D box edges. This contribution is followed by most image-based detection methods either in refinement step or as direct calculation on 3D objects \cite{li2019stereo,brazil2019m3drpn}. All we know in this constraint is that certain 3D points are projected onto 2D edges, but the corresponding relationship and the exact location of the projection are not clear. Therefore, it needs to exhaustively enumerate $8^4=4096$ configurations to determine the final correspondence and can only provide four constraints, which is not sufficient for fully 3D representation in 9 parameters. It led to the need to estimate other prior information. Nevertheless, possible inaccuracies in the 2D bounding boxes may result in a grossly inaccurate solution with a small number of constraints. Therefore, most of these methods obtain more accurate 2D box through a two-stage detector, which is difficult to get real-time speed.\\
    \textbf{Keypoints in Monocular 3D Object Detection.} It is believed that the detection accuracy of occluded and truncated objects can be improved by deducing complete shapes from vehicle keypoints \cite{chabot2017deep,zeeshan2014cars,murthy2017reconstructing}. They represent the regular-shape vehicles as a wire-frame template , which is obtained from a large number of CAD models. To train the keypoint detection network, they need to re-label the data set and even use depth maps to enhance the detection capability. \cite{he2019mono3d++} is most related to our work, which also considers the wire-frame model as prior information. Furthermore, It jointly optimizes the 2D box, 2D keypoints, 3D orientation, scale hypotheses, shape hypotheses, and depth with four different networks. This has limitations in run time. In contrast to prior work, we reformulate the 3D detection as the coarse keypoints detection task. Instead of predicting the 3D box based on an off-the-shelf 2D detectors or other data generators, we build a network to predict 9 of 2D keypoints projected by vertexes and center of 3D bounding box while minimize the reprojection error to find an optimal result. 	
           \begin{figure*}[htb]
		\begin{center}
			\includegraphics[width=2.1\columnwidth]{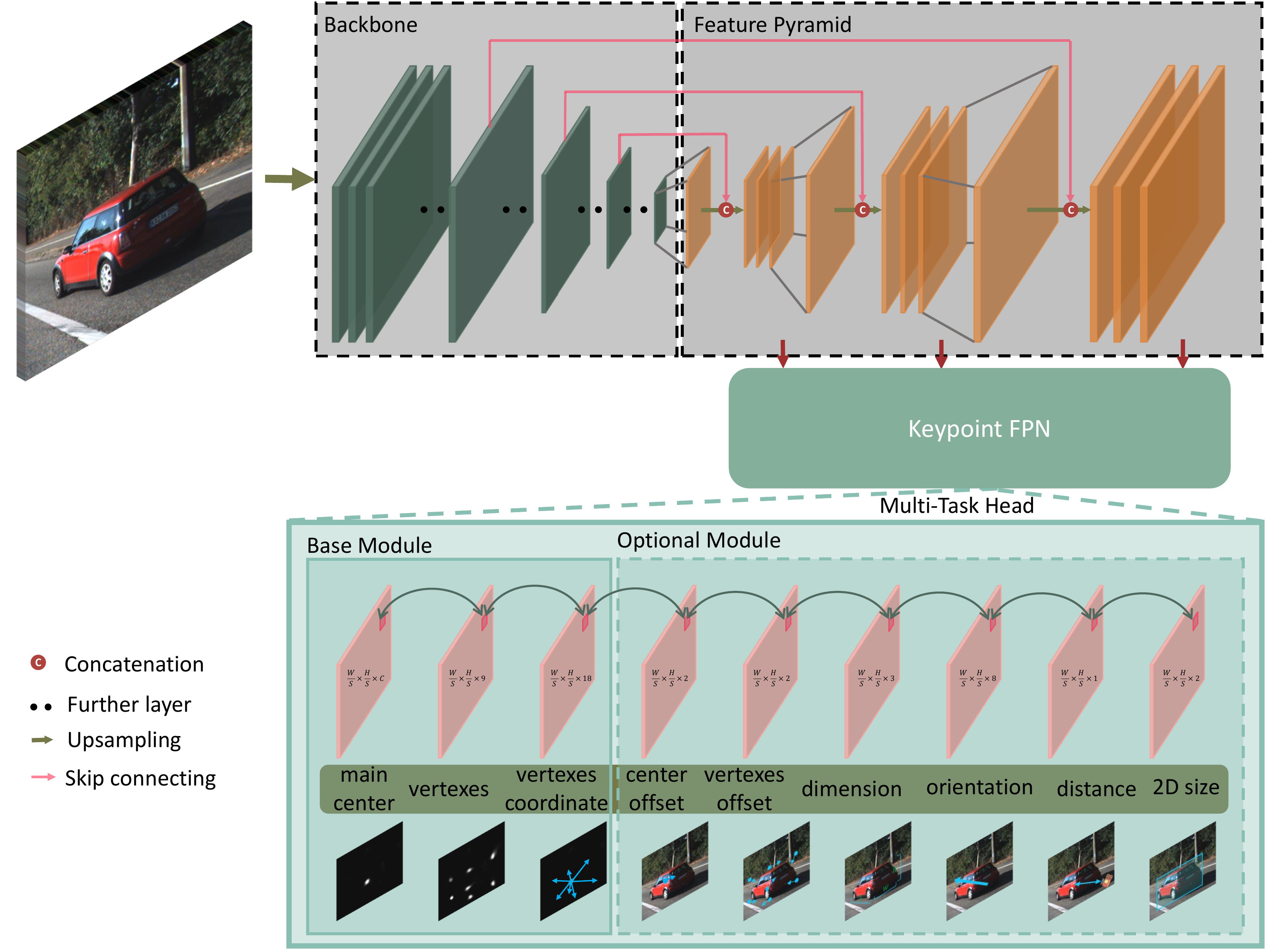}
		\end{center}
		 \caption{\textbf{An overview of proposed keypoint detection architecture:} It takes only the RGB images as the input and outputs main center heatmap, vertexes heatmap, and vertexes coordinate as the base module to estimate 3D bounding box. It can also predict other alternative priors to further improve the performance of 3D detection.}
		\label{fig:KDN}
	\end{figure*}
    \section{Proposed Method}\label{sec:proposed method}
    In this section. We first describe the overall architecture for keypoint detection. Then we detail how to find the 3D vehicles from the generated keypoints.
    \subsection{Keypoint Detection Network}
\begin{figure}
		\begin{center}
			\includegraphics[width=1\columnwidth]{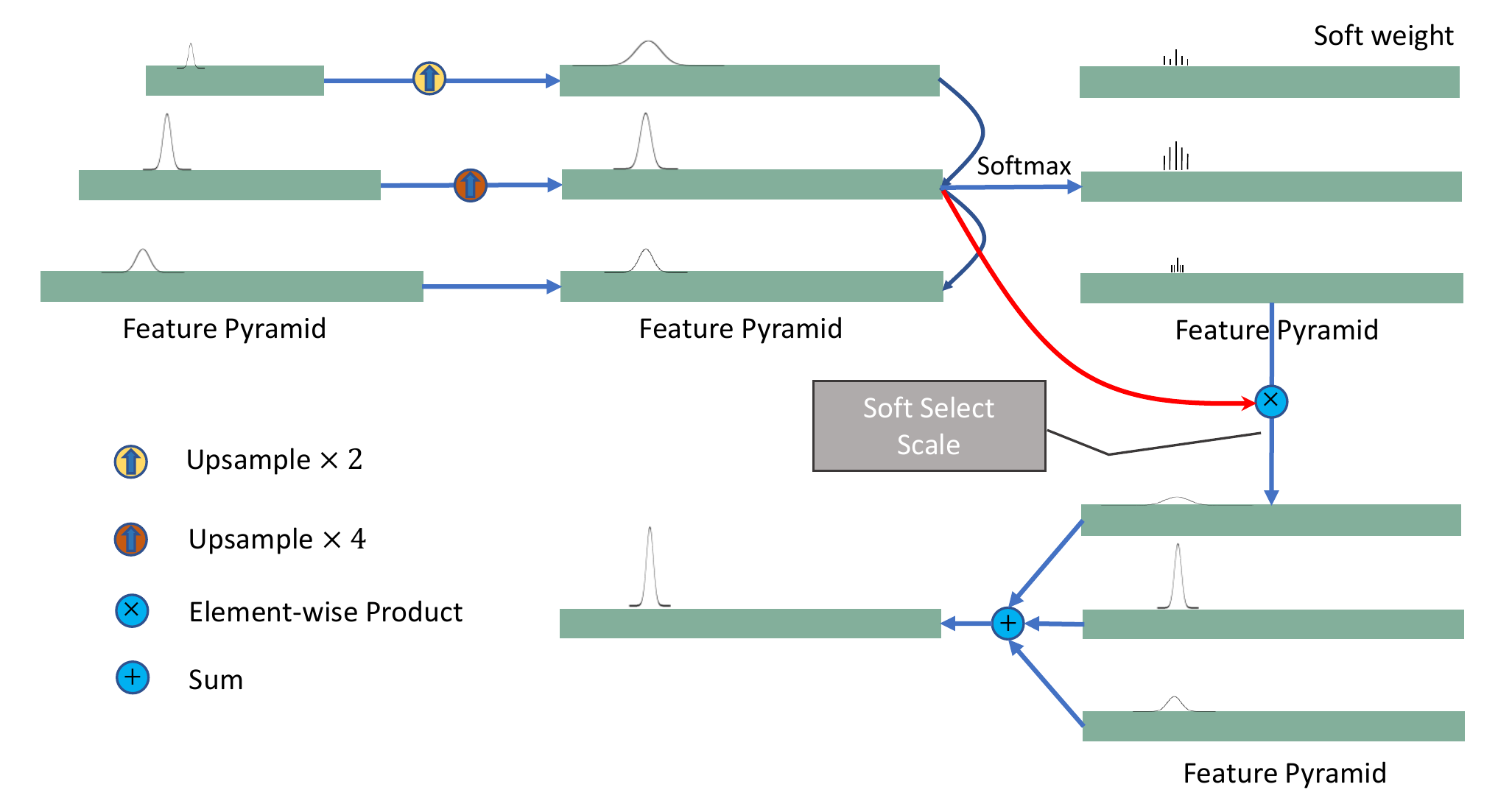}
		\end{center}
		 \caption{Illustration of our keypoint feature pyramid network(KFPN).}
		\label{fig:keypointnet}
	\end{figure}

    Our keypoint detection network takes an only RGB image as the input and generates the perspective points from vertexes and center of the 3D bounding box. As shown in Fig. \ref{fig:KDN}, it consists of three components: backbone, keypoint feature pyramid, and detection head. The main architecture adopts a one-stage strategy that shares a similar layout with the anchor-free 2D object detector \cite{tian2019fcos,kong2019foveabox,zhou2019objects,law2018cornernet}, which allows us to get a fast detection speed. Details of the network are given below.\\
    \textbf{Backbone.} For the trade-off between speed and accuracy, we use two different structures as our backbones: ResNet-18\cite{he2016deep} and DLA-34 \cite{yu2018deep}. All models take a single RGB image $I \in \mathbb{R}^{W \times H \times 3}$ and downsample the input with factor $S=4$. The ResNet-18 and DLA-34 build for image classification network, the maximal downsample factor is $\times 32$. We upsample the bottleneck thrice by three bilinear interpolations and $1 \times 1$ convolutional layer. Before the upsampling layers, we concatenate the corresponding feature maps of the low level while adding one $1\times1$ convolutional layers for channel dimension reduction. After three upsampling layers, the channels are 256, 128, 64, respectively.\\
    \textbf{Keypoint Feature Pyramid.} Keypoint in the image have no difference in size. Therefore, the keypoint detection is not suitable for using the Feature Pyramid Network(FPN) \cite{lin2017feature}, which detect multi-scale 2D box in different pyramid layers. We propose a novel approach Keypoint Feature Pyramid Network (KFPN) to detect scale-invariant keypoints in the point-wise space, as shown in Fig. \ref{fig:keypointnet}. Assuming we have $F$ scale feature maps, we first resize each scale $f, 1<f<F$ back to the size of maximal scale, which yields the feature maps $\hat{f}_{1<f<F}$. Then, we generate soft weight by a softmax operation to denote the importance of each scale. The finally scale-space score map $S_{score}$ is obtained by linear weighing sum. In detail, it can be defined as:
     \begin{equation}
	\label{eq:dimensionregression}
    \begin{array}{lr}
	\renewcommand*{\arraystretch}{1.5}
    \begin{aligned}
    S_{score}= \sum\limits_f{\hat{f} {\odot} softmax(\hat{f})}
    \end{aligned}
    \end{array}
	\end{equation}
    where $\odot$ denote element-wise product.
    \\
    \textbf{Detection Head.}
    The detection head is comprised of three fundamental components and six optional components which can be arbitrarily selected to boost the accuracy of 3D detection with a little computational consumption. Inspired by CenterNet \cite{zhou2019objects}, we take a keypoint as the maincenter for connecting all features. Since the 3D projection point of the object may exceed the image boundary in the case of truncation, the center point of the 2D box will be selected more appropriately. The heatmap can be define as $M \in [0,1]^{\frac{H}{S}\times {\frac{W}{S} \times C}}$, where $C$ is the number of object categories. Another fundamental component is the heatmap $V \in [0,1]^{\frac{H}{S}\times {\frac{W}{S} \times 9}}$ of nine perspective points projected by vertexes and center of 3D bounding box. For keypoints association of one object, we also regress an local offset $V_{c} \in R^{\frac{H}{S}\times {\frac{W}{S} \times 18}}$ from the maincenter as an indication. Keypoints of $V$ closest to the coordinates from $V_{c}$ are taken as a group of one object.

    Although the 18 constraints by the 9 keypoints have an ability to recover the 3D information of the object, more prior information can provide more constraints and further improve the detection performance. We offer a number of options to meet different needs for accuracy and speed. The center offset $M_{os} \in R^{\frac{H}{S}\times {\frac{W}{S} \times 2}}$ and vertexes offset $V_{os} \in R^{\frac{H}{S}\times {\frac{W}{S} \times 2}}$ are discretization error for each keypoint in heatmaps. The dimension $D \in \mathbb{R}^{\frac{H}{S}\times {\frac{W}{S} \times 3}}$ of 3D object have a smaller variance, which makes it easy to predict.
    The rotation $R(\theta)$ of an object only by parametrized by orientation $\theta$ (yaw) in the autonomous driving scene. We employ the \emph{Multi-Bin} based method \cite{mousavian20173d} to regress the local orientation. We classify the probability with cosin and sine offset of the local angle in one bin, which generates feature map of orientation $O \in \mathbb{R}^{\frac{H}{S}\times {\frac{W}{S} \times 8}}$ with two bins. We also regress the depth $Z \in \mathbb{R}^{\frac{H}{S}\times {\frac{W}{S} \times 1}}$ of 3D box center, which can be used as the initialization value to speed up the solution in Sec .\ref{sec:boxestimation}.\\
    \textbf{Training.} The all heatmaps of keypoint training strategy follow the \cite{zhou2019objects,law2018cornernet}. The loss solves the imbalance of positive and negative samples with focal loss \cite{lin2017focal}:
    \begin{equation}
	\label{eq:qualityloss}
    \begin{array}{lr}
	\renewcommand*{\arraystretch}{1.5}
	\quad \quad \quad \quad L_{kp}^{K} = -\frac{1}{N} \sum\limits_{k=1}^{K}\sum\limits_{x=1}^{H/S}\sum\limits_{y=1}^{W/S}\\
    \left\{
    \begin{aligned}
    (1-\hat{p}_{kxy})^\alpha log(\hat{p}_{kxy}) \quad \quad \quad    &if \quad p_{kxy}=1\\
    (1-p_{kxy})^\beta \hat{p}_{kxy} log(1-\hat{p}_{kxy}) \quad  &otherwise
    \end{aligned}
    \right.
    \end{array}
	\end{equation}
    where $K$ is the channels of different keypoints, $K=C$ in maincenter and $K=9$ in vertexes. $N$ is the number of maincenter or vertexes in an image, and $\alpha$ and $\beta$ are the hyper-parameters to reduce the loss weight of negative and easy positive samples. We set is $\alpha=2$ and $\beta=4$ in all experiments following \cite{zhou2019objects,law2018cornernet}. $p_{kxy}$ can be defined by Gaussian kernel $p_{xy}=exp\left(- \frac{x^2+y^2}{2 \sigma} \right )$ centered by ground truth keypoint $\tilde{p}_{xy}$. For $\sigma$, we find the max area $A_{max}$ and min area $A_{min}$ of 2D box in training data and set two hyper-parameters $\sigma_{max}$ and $\sigma_{min}$. We then define the $\sigma=A(\frac{\sigma_{max}-\sigma_{min}}{A_{max}-A_{min}})$ for a object with size $A$.
    For regression of dimension and distance, we define the residual term as:
 \begin{equation}
	\label{eq:dimensionregression}
    \begin{array}{lr}
	\renewcommand*{\arraystretch}{1.5}
    \begin{aligned}
	L_D=\frac{1}{3N}\sum\limits_{x=1}^{H/S}\sum\limits_{y=1}^{W/S}\mathbbm{1}_{xy}^{obj}\left( D_{xy}- \Delta \widetilde{D}_{xy}\right)^2\\
    L_{Z}=\frac{1}{N}\sum\limits_{x=1}^{H/S}\sum\limits_{y=1}^{W/S}\mathbbm{1}_{xy}^{obj}\left(log(Z_{xy})-log(\widetilde{Z}_{xy})\right)^2\\
    \end{aligned}
    \end{array}
	\end{equation}
   We set $\Delta \widetilde{D}_{xy}=log\frac{\widetilde{D}_{xy}-\bar {D}}{D_{\sigma}}$ , where $\bar {D}$ and $D_{\sigma}$ are the mean and standard deviation dimensions of training data. $\mathbbm{1}_{xy}^{obj}$ denotes if maincenter appears in position $x,y$.
   The offset of maincenter, vertexes are trained with an L1 loss following \cite{zhou2019objects}:
    \begin{equation}
	\label{eq:dimensionregression}
    \begin{array}{lr}
	\renewcommand*{\arraystretch}{1.5}
	L_{off}^{m}=\frac{1}{2N}\sum\limits_{x=1}^{H/S}\sum\limits_{y=1}^{W/S}\mathbbm{1}_{xy}^{obj}\left|M_{os}^{xy}-\left(\frac{p^m}{S}-\tilde{p}_{m} \right) \right|\\

    L_{off}^{v}=\frac{1}{2N}\sum\limits_{x=1}^{H/S}\sum\limits_{y=1}^{W/S}\mathbbm{1}_{xy}^{ver}\left|V_{os}^{xy}-\left(\frac{p^v}{S}-\tilde{p}_{v} \right) \right|
    \end{array}
	\end{equation}
    where $p^m, p^v$ are the position of maincenter and vertexes in the original image.
    The regression coordinate of vertexes with an L1 loss as:
    \begin{equation}
	\label{eq:dimensionregression}
    \begin{array}{lr}
	\renewcommand*{\arraystretch}{1.5}
    \begin{aligned}
	L_{ver}=\frac{1}{N}\sum\limits_{k=1}^{8}\sum\limits_{x=1}^{H/S}\sum\limits_{y=1}^{W/S}\mathbbm{1}_{xy}^{ver}\left|V_{c}^{(2k-1):(2k)xy}-\right.
\\
\left.\left|\frac{{p^v}-{p}^m}{S} \right| \right|
    \end{aligned}
    \end{array}
	\end{equation}
    Finial, we define the multi-task loss for keypoint detection as:
    \begin{equation}
	\label{eq:a}
    \begin{aligned}
	\renewcommand*{\arraystretch}{1.5}
	L=&\omega_{main} L_{kp}^C+\omega_{kpver} L_{kp}^8+\omega_{ver}L_{ver}+\omega_{dim} L_D+
    \\
    &\omega_{ori} L_{ori}+\omega_{Z} L_{dis}+\omega_{off}^{m} L_{off}^m+\omega_{off}^{v} L_{off}^v
    \end{aligned}
	\end{equation}
    We empirical set $\omega_{main}=1, \omega_{kpver}=1, \omega_{dim}=1, \omega_{ori}=0.5,\omega_{dis}=0.1, \omega_{off}^{m}=0.5$ and $\omega_{off}^{v}=0.5$ in our experimental.
    \subsection{3D Bounding Box Estimate}\label{sec:boxestimation}
    Consider an image $I$, a set of $i=1...N$ object are represented by 9 keypoints and other optional prior, given by our keypoint detection network. We define this keypoints as $\widehat{kp}_{ij}$ for $j\in1...9$, dimension as $\widehat{D}_i$, orientation as $\hat{\theta}_i$, and distance as $\widehat{Z}_i$. The corresponding 3D bounding box $B_i$ can be defined by its rotation $R_i(\theta) $, position $T_i = [T_i^x, T_i^y, T_i^z]^T $, and dimensions $D_i = [h_i, w_i, l_i]^T$. Our goal is to estimate the 3D bounding box $B_i$, whose projections of center and 3D vertexes on the image space best fit the corresponding 2D keypoints $\widehat{kp}_{ij}$. This can be solved by minimize the reprojection error of 3D keypoints and 2D keypoints. We formulate it and other prior errors as a nonlinear least squares optimization problem:
    \begin{equation}
	\label{eq:a}
    \begin{aligned}
	\renewcommand*{\arraystretch}{1.5}
	R^*, T^*, D^*=& \mathop{\arg\max}\limits_{\{R,T,D\}}\sum\limits_{R_i,T_i,D_i}\left\|e_{cp}\left(R_i,T_i,D_i,\widehat{kp}_i\right)\right\|^2_{\Sigma_{i}}
    \\
    +&\omega_d\left\|e_d\left(D_i,\widehat{D}_i\right)\right\|^2_2 +\omega_r\left\|e_r\left(R_i,\hat{\theta}_i\right)\right\|^2_2
    \end{aligned}
	\end{equation}
    where $e_{cp}(..), e_d(..), e_r(..)$ are measurement error of camera-point, dimension prior and orientation prior respectively. We set $\omega_d=1$ and $\omega_r=1$ in our experimental.
    $\Sigma$ is the covariance matrix of keypoints projection error. It is the confidence extracted from the heatmap corresponding to the keypoints:
    \begin{equation}
	\Sigma_i=diag\left(Softmax\left(V(\widehat{kp}_i^{1:8}),M(\widehat{kp}_i^{9})\right)\right)
    \end{equation}
    In the rest of the section, we will first define this error item, and then introduce the way to optimize the formulation.\\
    \textbf{Camera-Point.} Following the \cite{geiger2012we}, we define the homogeneous coordinate of eight vertexes and 3D center as:
    \begin{equation}
    \begin{split}
    P_{3D}^i=diag(D_i)Cor \quad \quad \quad \quad \quad \quad\\
    Cor=
    \left[
    \begin{smallmatrix}
    0   & 0   & 0    &  0 &-1  &  -1 &  -1 &  -1 & -1/2  \\
    1/2 & -1/2& -1/2 &1/2 &1/2 &-1/2 &-1/2 &1/2  &  0   \\
    1/2 & 1/2 & -1/2 &-1/2&1/2 &1/2  &-1/2 &-1/2 &  0\\
    1   & 1   &1     & 1  &1   &1    &1    &1    &  1
    \end{smallmatrix}
    \right]\\
    \end{split}
    \end{equation}
   Given the camera intrinsics matrix $K$, the projection of these 3D points into the image coordinate is:
    \begin{equation}
	\label{eq:a}
    \begin{aligned}
	\renewcommand*{\arraystretch}{1.5}
    {kp_i}&=\frac{1}{s_i}K
    \left[
    \begin{matrix}
    R& T\\
    0^T&1
    \end{matrix}
    \right]diag(D_i)Cor\\
    &=\frac{1}{s_i}K\exp(\xi^{\wedge})diag(D_i)Cor
    \end{aligned}
	\end{equation}
    where $\xi \in \mathfrak{se}_3$ and $\exp$ maps the $\mathfrak{se}_3 $ into $SE_3$ space. The projection coordinate should fit tightly into 2D keypoints detected by the detection network. Therefore, the camera-point error is then defined as:
    \begin{equation}
    \begin{aligned}
	\renewcommand*{\arraystretch}{1.5}
    e_{cp}=\widehat{kp}_i-{kp_i}
    \end{aligned}
	\end{equation}
    Minimizing the camera-point error needs the Jacobians in $\mathfrak{se}_3$ space. It is given by:
    \begin{equation}
    \begin{aligned}
	\renewcommand*{\arraystretch}{1.5}
    &\frac{\partial e_{cp}}{\partial \delta \xi}=-
    \left[
    \begin{matrix}
    \frac{f_x}{Z^{'}}& 0& -\frac{f_xX^{'}}{Z^{'^2}}\\
    0&\frac{f_y}{Z^{'}} 0&-\frac{f_yY^{'}}{Z^{'^2}}
    \end{matrix}
    \right] \cdot
     \left[
    \begin{matrix}
    I,& -P^{'^\wedge}
    \end{matrix}
    \right]\\
    &\frac{\partial e_{cp}}{\partial D_i}=- \frac{1}{9}\sum\limits_{col=1}^{9}
    \left[
    \begin{matrix}
    \frac{f_x}{Z^{'}}& 0& -\frac{f_xX^{'}}{Z^{'^2}}\\
    0&\frac{f_y}{Z^{'}} 0&-\frac{f_yY^{'}}{Z^{'^2}}
    \end{matrix}
    \right]\cdot R \cdot Cor_{col}
    \end{aligned}
	\end{equation}
    where $P^{'}=[X^{'},Y^{'},Z^{'}]^{T}=\left(\exp(\xi^{\wedge}P)\right)_{1:3}$.\\
    \textbf{Dimension-Prior:} The $e_d$ is sample defined as:
    \begin{equation}
    \begin{aligned}
	\renewcommand*{\arraystretch}{1.5}
    e_{d}=\widehat{D}_i-D_i
    \end{aligned}
	\end{equation}
    \textbf{Rotation-Prior:} We define $e_r$ in $SE3$ space and use $log$ to map the error into its tangent vector space:
    \begin{equation}
    \begin{aligned}
	\renewcommand*{\arraystretch}{1.5}
    e_{r}=\log(R^{-1}R(\hat{\theta}))^{\vee}_{\mathfrak{se}_3}
    \end{aligned}
	\end{equation}
     These multivariate equations can be solved via the Gauss-newton or Levenberg-Marquardt algorithm in the g2o library \cite{kummerle2011g}. A good initialisation is mandatory using this optimization strategy. We adopt the prior information generated by keypoint detection network as the initialization value, which is very important in improving the detection speed.
    \section{Experimental}\label{sec:experimental}
    \subsection{Implementation Details}
      \begin{table*}[!t]
\footnotesize
\begin{center}
\begin{tabular}{| p{2.9cm}<{\centering} | p{1.4cm}<{\centering}  | c | c@{/}c | c@{/}c | c@{/}c || c@{/}c | c@{/}c | c@{/}c |}
\hline
\multirow{2}{*}{Method} & \multirow{2}{*}{Extra} & \multirow{2}{*}{Time} & \multicolumn{6}{c||}{$\text{AP}_{3D}$ (IoU=0.5)} & \multicolumn{6}{c|}{$\text{AP}_{3D}$ (IoU=0.7)} \\
\cline{4-15}
& & & \multicolumn{2}{c|}{Easy} & \multicolumn{2}{c|}{Moderate} & \multicolumn{2}{c||}{Hard} & \multicolumn{2}{c|}{Easy} & \multicolumn{2}{c|}{Moderate} & \multicolumn{2}{c|}{Hard} \\
\hline
Mono3D \cite{chen2016monocular} & Mask & 4.2 s & 25.19 & - & 18.20 & - & 15.52 & - & 2.53 & - & 2.31 & - & 2.31&- \\
3DOP \cite{chen20173d} & Stereo & 3 s &   46.04 & - &  34.63 & - &  30.09 & - & 6.55 & - & 5.07 & - & 4.10 &-\\
MF3D \cite{xu2018multi} & Depth & - & 47.88 & 45.57 & 29.48 & 30.03 & 26.44 & 23.95 & 10.53 & 7.85 & 5.69 & 5.39  & 5.39 & 4.73 \\
Mono3D++ \cite{he2019mono3d++} & Depth+Shape & $>$0.6s & 42.00 & - & 29.80 & - & 24.20 & - & 10.60 & - & 7.90 & - & 5.70 & - \\
\hline
\hline
GS3D \cite{Li_2019_CVPR} & None &2.3s& 32.15 & 30.60 & 29.89 & 26.40 & 26.19 & 22.89 & 13.46 & 11.63 & 10.97 & 10.51 & 10.38  & 10.51  \\
M3D-RPN \cite{brazil2019m3drpn} & None &0.16s& 48.96 & 49.89 & 39.57 & 36.14 & 33.01 & 28.98 & 20.27 & \textbf{20.40} & \textbf{17.06} & \textbf{16.48} & 15.21 & 13.34 \\
Deep3DBox \cite{mousavian20173d} & None & - & 27.04 & - & 20.55 & - & 15.88 & - & 5.85 & - & 4.10 & - & 3.84 & - \\
MonoGRNet \cite{qin2019monogrnet} & None & 0.06s & 50.51 & - & 36.97 & - & 30.82 & - & 13.88 & - & 10.19 & - & 7.62 & - \\
FQNet\cite{liu2019deep}& None & 3.33s & 28.16 & 28.98 & 21.02 & 20.71 & 19.91 & 18.59 & 5.98 & 5.45 & 5.50 & 5.11 & 4.75 & 4.45 \\
\hline
Ours (ResNet18)& None & {0.035} s & 47.43  & 46.52 & 33.86  & 32.61 & 31.04 & 30.95 &   18.13& 18.38 & 14.14 & 14.66 & 13.33 & 12.35 \\
Ours (DLA34)& None & {0.055} s & \textbf{54.36}  & \textbf{52.59} & \textbf{41.90}  & \textbf{40.96} & \textbf{35.84} & \textbf{34.95} & \textbf{20.77}& 19.47 & 16.86 & 16.29 & \textbf{16.63} &
\textbf{15.57} \\
\hline
\end{tabular}
\vspace{1.5mm}
\caption{Comparison of our method to image-based 3D detection frameworks for car category evaluated using metric $AP_{3D}$ on $val_1$ / $val_2$ of KITTI data set. ¡°Extra¡± means the extra data used in training.}
\label{tab:3d}
\end{center}
\end{table*}
\begin{table*}[!t]
\footnotesize
\begin{center}
\begin{tabular}{| c | c | c | c@{/}c | c@{/}c | c@{/}c || c@{/}c | c@{/}c | c@{/}c |}
\hline
\multirow{2}{*}{Method} & \multirow{2}{*}{Extra} & \multirow{2}{*}{Time} & \multicolumn{6}{c||}{$\text{AP}_{BEV}$ (IoU=0.5)} & \multicolumn{6}{c|}{$\text{AP}_{BEV}$ (IoU=0.7)} \\
\cline{4-15}
& & & \multicolumn{2}{c|}{Easy} & \multicolumn{2}{c|}{Moderate} & \multicolumn{2}{c||}{Hard} & \multicolumn{2}{c|}{Easy} & \multicolumn{2}{c|}{Moderate} & \multicolumn{2}{c|}{Hard} \\
\hline
Mono3D \cite{chen2016monocular} & Mask & 4.2 s & 30.50 & - & 22.39& - & 19.16 & -  & 5.22 & - & 5.19 &- & 4.13 & - \\
3DOP \cite{chen20173d} & Stereo & 3 s &  55.04 & - &  41.25 & - &  34.55 & - & 12.63 & - & 9.49 & - & 7.59 & - \\
MF3D \cite{xu2018multi} & Depth & - & 55.02 & 54.18 & 36.73 & 38.06 & 31.27 & 31.46 & 22.03 & 19.20 & 13.63 & 12.17  & 11.60 & 10.89 \\
Mono3D++ \cite{he2019mono3d++} & Depth+Shape & $>$0.6s & 46.70 & - & 34.30 & - & 28.10 & - & 16.70 & - & 11.50 & - & 10.10 & - \\
\hline
\hline
GS3D \cite{Li_2019_CVPR} & None &2.3s& - &- &- &- &- &- &- &- &- &- &- &-  \\
M3D-RPN \cite{brazil2019m3drpn} & None &0.16s& 55.37 & 55.87 & 42.49 & 41.36 & 35.29 & 34.08 & \textbf{25.94} & \textbf{26.86} & 21.18 & 21.15 & 17.90 & 17.14 \\
Deep3DBox \cite{mousavian20173d} & None & - & 30.02 & - & 23.77 & - & 18.83 & - & 9.99 & - & 7.71 & - & 5.30 & - \\
MonoGRNet \cite{qin2019monogrnet} & None & 0.06s & - &- &- &- &- &- &- &- &- &- &- &- \\
FQNet\cite{liu2019deep}& None & 3.33s & 32.57 & 33.37 & 24.60 & 26.29 & 21.25 & 21.57& 9.50 & 10.45 & 8.02 & 8.59 & 7.71 & 7.43 \\
\hline
Ours(ResNet18) & None & {0.035}s & 52.79  & 41.91 & 35.92  & 34.28 & 33.02 & 28.88 & 20.81 & 10.84 & 16.60 & 16.48 & 15.80 & 15.45 \\
Ours (DLA34)& None & {0.055} s & \textbf{57.47}  & \textbf{56.90} & \textbf{44.16}  & \textbf{44.69} & \textbf{42.31} & \textbf{41.75} & 25.56& 24.74 & \textbf{22.12} & \textbf{22.03} & \textbf{20.91} &
\textbf{18.05} \\
\hline
\end{tabular}
\vspace{1.5mm}
\caption{ Comparison of our method to image-based 3D detection frameworks for car category, evaluated using metric $AP_{BEV}$ on $val_1$ / $val_2$ of KITTI data set.}
\label{tab:BEV}
\end{center}
\end{table*}
    We evaluated our experiments on the KITTI 3D detection benchmark \cite{geiger2012we}, which has a total of 7481 training images and 7518 test images.
    We follow the \cite{chen20173d} and \cite{xiang2017subcategory-aware} to split the training set as $train1,val_1$ and $train2,val_2$ respectively. We comprehensively compare our framework and other method on this two validation as well as test set.

    We implemented our network using PyTorch, with the machine i7-8086K CPU and 2 1080Ti GPUs. We pad the original image to $1280\times384$ for training and testing.
    We project the 3D bounding box of Ground Truths in the left and right images to obtain Ground Truth keypoints and use the flipping as image augmentation, which makes our dataset is quadruple with the origin training set. We run Adam \cite{kingma2014adam:} optimizer with a base learning rate of 0.0002 for 300 epochs and reduce $10\times$ at 150 and 180 epochs. For standard deviation of Gaussian kernel, we set $\sigma_{max}=19$ and $\sigma_{min}=3$. Based on the statistics of KITTI dataset, we set $\tilde{l}=3.89, \tilde{w}=1.62, \tilde{h}=1.53$ and $\sigma_{\tilde {l}}=0.41,\sigma_{\tilde {w}}=0.1,\sigma_{\tilde {h}}=0.13$. In the inference step, after $3\times 3$  max pooling, we filter the maincenter and keypoints with threshold 0.4 and 0.1 respectively, and only keypoints that in the image size range is sent into the geometric constraint module. The backbone networks are initialized by a classification model pre-trained on the ImageNet data set. Finally, The ResNet-18 takes about three days with batch size 16 and DLA-34 for four days with batch size 30 in training.
    \subsection{Comparison with Other Methods}
    To fully evaluate the performance of our keypoint-based method, for each task three official evaluation metrics be reported in KITTI: average precision for 3D intersection-over-union ($AP_{3D}$), average precision for Birds Eye View ($AP_{BEV}$), and Average Orientation Similarity (AOS) if 2D bounding box available. We evaluate our method at three difficulty settings: easy, moderate, and hard, according to the object's occlusion, truncation, and height in the image space \cite{geiger2012we}. \\     $\bm{AP_{3D}}$ \textbf{and} $\bm{AP_{BEV}}$. We compare our method with current image-based SOTA approaches and also provide a comparison about running time. However, it is not realistic to list the running times of all previous methods because most of them do not report their efficiency. The results $AP_{3D}$, $AP_{BEV}$ and running time are shown in Table \ref{tab:3d} and \ref{tab:BEV}, respectively. ResNet-18 as the backbone achieves the best speed while our accuracy outperforms most of the image-only method. In particular, it is more than 100 times faster than Mono3D \cite{chen2016monocular} while outperforms over 10\% for both $AP_{BEV}$ and $AP_{3d}$ across all datasets. In addition, our ResNet-18 method is more than 75 times faster while having a comparable accuracy than 3DOP \cite{chen20173d}, which employs stereo images as the input. DLA-34 as the backbone achieves the best accuracy while having relatively good speed. It is faster about 3 times than the recently proposed M3D-RPN \cite{brazil2019m3drpn} while achieves the improvement in most of the metrics. Note that comparing our method with this all approaches is unfair because most of these approaches rely on extra stand-alone network or data in addition to monocular images. Nevertheless, we achieve the best speed with better performance.\\
    \textbf{Results on the KITTI testing set.} We also evaluate our results on the KITTI testing set, as shown in Table. \ref{tab:testset}. More details can be found on the KITTI website \footnote{\url{http://www.cvlibs.net/datasets/kitti/eval_object.php?obj_benchmark=3d}}.
    \begin{table}[!ht]
\setlength{\tabcolsep}{1mm}
\footnotesize
\begin{center}
\begin{tabular}{| c | c | c | c | c |}
\hline
\multirow{2}{*}{Method} &\multirow{2}{*}{time} & \multicolumn{3}{c|}{AP$_\text{3D}$(IoU=0.7)} \\
\cline{3-5}
&&Easy & Mode & Hard \\
\hline
GS3D\cite{Li_2019_CVPR}&2.3s& 7.69 & 6.29 & 6.16\\
\hline
MonoGRNet\cite{qin2019monogrnet}&0.06s& 9.61 & 5.74 & 4.25\\
\hline
M3D-RPN\cite{brazil2019m3drpn}&0.16s&\textbf{14.76} & 9.71 & 7.42\\
\hline
FQNet\cite{liu2019deep}&3.33s&2.77 & 1.51 & 1.01\\
\hline
Ours(DLA34)&0.055s& 13.61 & \textbf{10.09}  & \textbf{8.18} \\
\hline
\end{tabular}
\vspace{1mm}
\caption{Comparing 3D detection $AP_{3D}$ on KITTI testing set. We use the DLA-34 as the backbone.}
\label{tab:testset}
\end{center}
\end{table}\\
    \subsection{Qualitative Results}
\begin{figure*}
		\begin{center}
			\includegraphics[width=2.1\columnwidth]{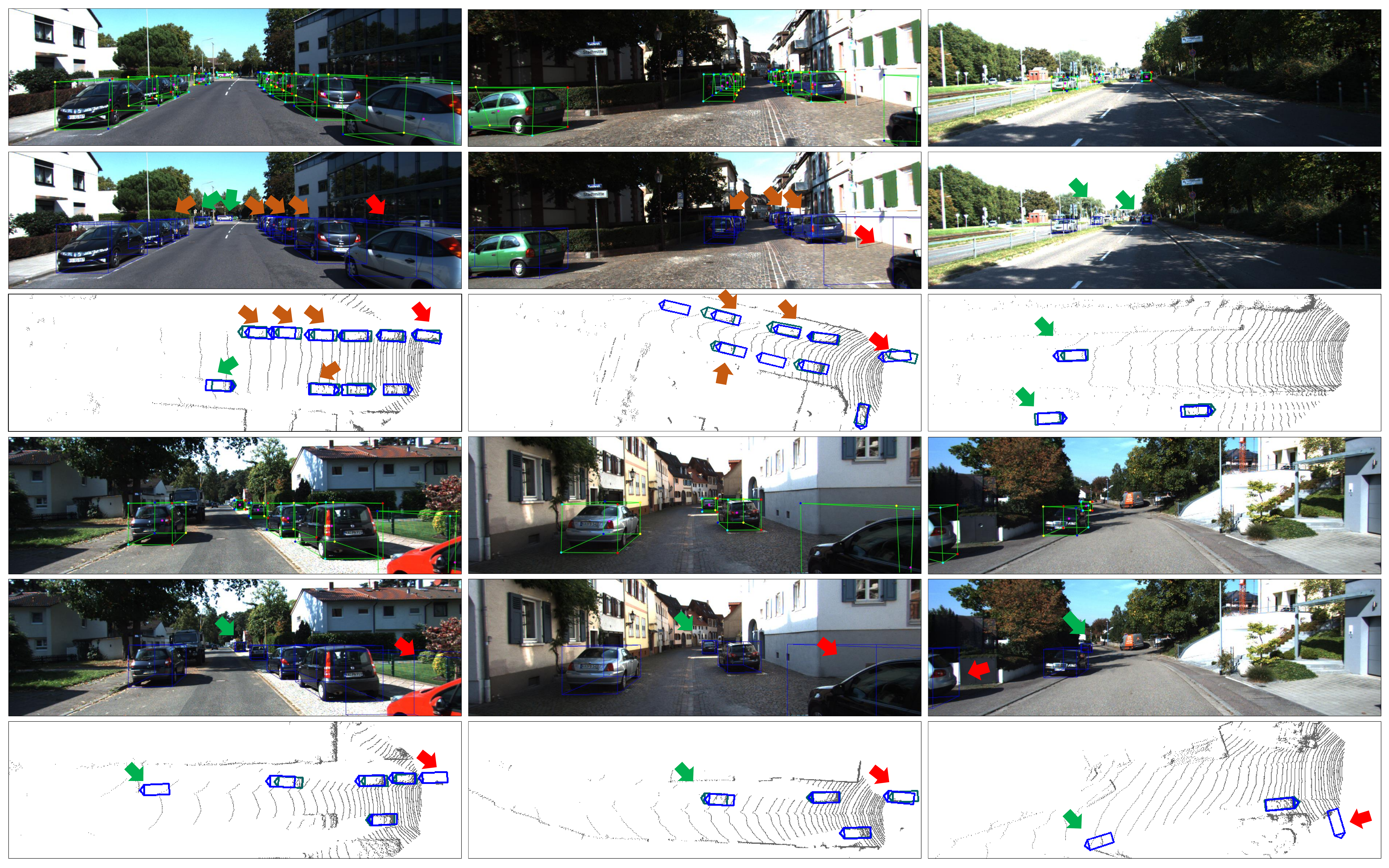}
		\end{center}
		\caption{Qualitative results of our 3D detection. From top to bottom are keypoints, projections of the 3D bounding box and bird's eye view image, ground truths in green and our results in blue. The crimson arrows, green arrows, and red arrows point to occluded, distant, and truncated objects, respectively.}
		\label{fig:qualitative}
\end{figure*}

    Fig. \ref{fig:qualitative} shows some qualitative results of our method. We visualize the keypoint detection network outputs, geometric constraint module outputs and BEV images.
    The results of the projected 3D box on image demonstrate than our method can handle crowded and truncated objects. The results of the BEV image show that our method has an accuracy localization in different scenes.
    \subsection{Ablation Study}
    \textbf{Effect of Optional Components.} Three optional components be employed to enhance our method: dimension, orientation, distance and keypoints offset. We experiment with different combinations to demonstrate their effect on 3D detection. The results are shown in Table.\ref{tab:opcom}, we train our network with DLA-34 backbone and evaluate it using $AP_{3D}$ and $AP_{BEV}$. The combinations of dimension, orientation, distance and keypoints offset achieve the best accuracy meanwhile have a faster running speed. This is because we take the output predicted by our network as the initial value of the geometric optimization module, which can reduce the search space of the gradient descent method.
    \begin{table}[!ht]
\setlength{\tabcolsep}{1mm}
\footnotesize
\begin{center}
\begin{tabular}{| c | c  | c |c|c| c | c | c || c | c | c |}
\hline
\multirow{2}{*}{dim} &\multirow{2}{*}{ori}&\multirow{2}{*}{dist}&\multirow{2}{*}{off}&\multirow{2}{*}{time(s)} & \multicolumn{3}{c||}{AP$_\text{3D}$(IoU=0.5)}&\multicolumn{3}{c|}{AP$_{\text{3D}}$(IoU=0.7)} \\
\cline{6-11}
&&                      &&&Easy & Mode & Hard & Easy & Mode & Hard\\
\hline
$\surd$&&               &&0.058&51.21 & 40.73 & 35.00 &18.23 & 17.05 & 15.94\\
\hline
&$\surd$&               &&0.061&25.35 & 22.33 & 21.18 & 3.12 & 3.43 & 2.97\\
\hline
$\surd$&$\surd$&        &&0.057&54.18 & 41.34 & 34.89 &20.23 & 16.02 & 15.94\\
\hline
$\surd$&$\surd$&$\surd$ &&0.055&54.20& 41.56 & 35.13&20.76 & 16.80 & 16.25\\
\hline
$\surd$&$\surd$&$\surd$ &$\surd$&0.055&54.36 & 41.90 & 35.84&20.77 & 16.86 & 16.36\\
\hline

\end{tabular}
\vspace{1mm}
\caption{Ablation study of different optional selecting results on $val_1$ set. We use the DLA-34 as the backbone.}
\label{tab:opcom}
\end{center}
\end{table}\\
\textbf{Effect of Keypoint FPN.}
We propose keypoint FPN as a strategy to improve the performance of multi-scale keypoint detection. To better understand its effect, we compare the $AP_{3D}$ and $AP_{BEV}$ with and without KFPN. The details are shown in Table. \ref{tab:kfpncom}, using KFPN achieves the improvement across all sets while no significant
change in time consumption.
\begin{table}[!ht]
\setlength{\tabcolsep}{1mm}
\footnotesize
\begin{center}
\begin{tabular}{| c | c | c | c | c || c | c | c |}
\hline
\multirow{2}{*}{KFPN} &\multirow{2}{*}{time} & \multicolumn{3}{c||}{AP$_\text{3D}$(IoU=0.7)}&\multicolumn{3}{c|}{AP$_{\text{3D}}$(IoU=0.5)} \\
\cline{3-8}
&&Easy & Mode & Hard & Easy & Mode & Hard\\
\hline
w/o&0.054& 50.14 & 40.73 & 34.94 & 17.47 & 15.99&15.36\\
\hline
w/& 0.055&54.36 & 41.90 & 35.84&20.77 & 16.86 & 16.36\\
\hline
\end{tabular}
\vspace{1mm}
\caption{Comparing 3D detection $AP_{3D}$ of w/o KFPN and w/ KFPN for car category on $val_1$ set. We use the DLA-34 as the backbone.}
\label{tab:kfpncom}
\end{center}
\end{table}\\
\textbf{2D Detection and Orientation.}
Although our focus is on 3D detection, we also compare the performance of our methods in 2D detection and orientation evaluation. We report the AOS and AP with a threshold IoU=0.7 for comparison. The results are shown in Table. \ref{tab:2D}, the Deep3DBox train MS-CNN \cite{cai2016unified} in KITTI to produce 2D bounding box and adopt VGG16 \cite{simonyan2014very} for orientation prediction, which gives him the highest accuracy. Deep3Dbox takes advantage of better 2D detectors, however, our $AP_{3D}$ outperforms it by about 20\% in moderate sets, which emphasize the importance of customizing the network specifically for 3D detection. Another interesting finding is that the 2D accuracy of back-projection 3D results is better than the direct prediction, thanks to our method that can infer the occlusive area of the object.
\begin{table}[!ht]
\begin{center}
\begin{tabular}{| c | c@{/}c | c@{/}c |}
\hline
Method & \multicolumn{2}{c|}{$\text{AP}_\text{2D}$} & \multicolumn{2}{c|}{$\text{AOS}$} \\
\hline
Mono3D \cite{chen2016monocular}  & 88.67 &- & 86.28 &-  \\
3DOP \cite{chen20173d} & 88.07 &-  & 85.80&-  \\
Deep3DBox \cite{mousavian20173d} & -& \textbf{97.20} &- & \textbf{96.68}   \\
DeepMANTA \cite{chabot2017deep} & \textbf{90.89}  &91.01 & \textbf{90.66}& 90.66 \\
GS3D \cite{Li_2019_CVPR}& 88.85& 90.02& 87.52 & 89.13 \\
\hline
Ours(2D) & 90.14 & 91.85 & 89.58 & 89.22 \\
Ours(3D) & 90.41 & 92.08 & 89.95 & 89.40 \\
\hline
\end{tabular}
\vspace{1.5mm}
\caption{Comparing of 2D detection $AP_{2D}$ with IoU=0.7 and orientation AOS results for car category evaluated on $val_1$ / $val_2$ of KITTI data set. Only the results under the moderate criteria are shown. Ours(2D) represents the results from the keypoint detection network, and Ours(3D) is the 2D bounding box of the projected 3D box. }
\label{tab:2D}
\end{center}
\end{table}

    \section{Conclusion}\label{sec:conclusion}
    In this paper, we have proposed a faster and more accurate monocular 3D object detection method for autonomous driving scenarios. We reformulate 3D detection as the keypoint detection problem and show how to recover the 3D bounding box by using keypoints and geometric constraints. We specially customize the point detection network for 3D detection, which can output keypoints of the 3D box and other prior information of the object using only images. Our geometry module formulates this prior to easy-to-optimize loss functions. Our approach generates a stable and accurate 3D bounding box without containing stand-alone networks, additional annotation while achieving real-time running speed.

{\small
\bibliographystyle{ieee_fullname}
\bibliography{ref}
}

\end{document}